%% file: emnlp2020.tex
\documentclass[11pt,a4paper]{article}
\usepackage[hyphenbreaks]{breakurl}
\usepackage[hyperref]{emnlp2020}
\usepackage[utf8]{inputenc}
\usepackage{times}
\usepackage{latexsym}

\usepackage{url}
\usepackage{amsmath}
\usepackage{amsfonts}
\usepackage{xfrac}
\usepackage{mathtools}
\usepackage[table]{colortbl}%
\usepackage{tikz}
\usepackage{caption}
\usepackage{booktabs}
\usepackage{adjustbox}
\usepackage{xspace}
\usepackage[normalem]{ulem}
\usepackage{subcaption}
\usepackage{pifont}
\usepackage{bbm}
\usepackage{xcolor}
\usepackage{algorithm}
\usepackage{multirow}
\usepackage[noend]{algpseudocode}

\usepackage[disable]{todonotes}

\usepackage{cleveref}
\crefname{section}{\S}{\S\S}
\crefname{table}{Tab.}{}
\crefname{figure}{Fig.}{}
\crefname{algorithm}{Alg.}{}
\crefname{equation}{Eq.}{}
\crefname{appendix}{App.}{}
\crefformat{section}{\S#2#1#3}  %

\DeclareUnicodeCharacter{01B0}{u}%
\DeclareUnicodeCharacter{01A1}{o}%
\DeclareUnicodeCharacter{0300}{r}%

\newcommand{\src}{$\mathbf{S}$\space}
\newcommand{\trg}{$\mathbf{T}$\space}
\definecolor{worse}{RGB}{245, 121, 58}
\newcommand{\worse}{\cellcolor{worse}}
\definecolor{better}{RGB}{133, 192, 249}
\newcommand{\better}{\cellcolor{better}}
\newcommand\blfootnote[1]{%
  \begingroup
  \renewcommand\thefootnote{}\footnote{#1}%
  \addtocounter{footnote}{-1}%
  \endgroup
}

\usepackage{microtype}

\aclfinalcopy %

\input{table}

\title{Do Explicit Alignments Robustly Improve Multilingual Encoders?}

\author{Shijie Wu \and Mark Dredze \\
Department of Computer Science \\
Johns Hopkins University \\
{\tt shijie.wu@jhu.edu, mdredze@cs.jhu.edu}
}

\date{}

\begin{document}
\maketitle
\begin{abstract}
Multilingual BERT \citep[mBERT]{devlin-etal-2019-bert},
XLM-RoBERTa \citep[XLMR]{conneau2019unsupervised} and other unsupervised multilingual encoders
can effectively learn cross-lingual representation.
Explicit alignment objectives based on bitexts like Europarl or MultiUN have been shown to further improve these representations.
However, word-level alignments are often suboptimal and such bitexts are unavailable for many languages.
In this paper, we propose a new contrastive alignment objective that can better utilize such signal,
and examine whether these previous alignment methods can be adapted to noisier sources of aligned data: a randomly sampled 1 million pair subset of the OPUS collection.
Additionally, rather than report results on a single dataset with a single model run, we report the mean and standard derivation of multiple runs with different seeds, on four datasets and tasks.
Our more extensive analysis finds that, while our new objective outperforms previous work, overall these methods do not improve performance with a more robust evaluation framework.
Furthermore, the gains from using a better underlying model eclipse any benefits from alignment training. 
These negative results dictate more care in evaluating these methods and suggest limitations in applying explicit alignment objectives.
\blfootnote{Code is available at \url{https://github.com/shijie-wu/crosslingual-nlp}.}
\end{abstract}

\section{Introduction}

Unsupervised massively multilingual encoders including multilingual BERT \cite[mBERT]{devlin-etal-2019-bert} and XLM-RoBERTa \citep[XLMR]{conneau2019unsupervised} are now standard tools for zero-shot cross-lingual transfer for NLP tasks \citep{wu-dredze-2019-beto,Xia2020WhichA}.
While almost all encoders are pretrained without explicit cross-lingual objective, i.e. enforcing similar words from different languages have similar representation, improvements can be attained through the use of explicit cross-lingually linked data during pretraining, such as bitexts \citep{lample2019cross,huang-etal-2019-unicoder,ji2019cross} and dictionaries \citep{wu2019emerging}.
As with cross-lingual embeddings \citep{ruder2019survey}, these data can be used to support explicit alignment objectives with either linear mappings \citep{wang-etal-2019-cross,wang2019cross,wu2019emerging,liu-etal-2019-investigating} or fine-tuning \citep{cao2020multilingual}.

However, as word-level alignments from an unsupervised aligner are often suboptimal, we develop a new cross-lingual alignment objective for training our model.
We base on our objective on contrastive learning, in which two similar inputs -- such as from a bitext -- are directly optimized to be similar, relative to a negative set. These methods have been effective in computer vision tasks \citep{he2019momentum,chen2020simple}.
Additionally, most previous work on contextual alignments consider high-quality bitext like Europarl \citep{koehn2005europarl} or MultiUN \citep{eisele-chen-2010-multiun}. While helpful, these resources are unavailable for most languages for which we seek a zero-shot transfer.
To better reflect the quality of bitext available for most languages, we additionally use OPUS-100 \citep{zhang2020improving}, a randomly sampled 1 million subset (per language pair) of the OPUS collection \citep{tiedemann-2012-parallel}.

We show that our new contrastive learning alignment objectives outperform previous work \citep{cao2020multilingual} when applied to bitext from previous works or the OPUS-100 bitext. However, our experiments also produce a negative result. While previous work showed improvements from alignment-based objectives on zero-shot cross-lingual transfer for a single task (XNLI) with a single random seed, our more extensive analysis tells a different story. We report the mean and standard derivation of multiple runs with the same hyperparameters and different random seeds. We find that previously reported improvements disappear, even while our new method shows a small improvement. Furthermore, we extend the evaluation to multiple languages on 4 tasks, further supporting our conclusions. 
Finally, we evaluate XLMR$_{\text{large}}$ on these tasks, which dominate the results obtained from the alignment objectives. 
We conclude that explicit alignments do not improve cross-lingual representations under a more extensive evaluation with noisier bitexts, and improvements are lost when compared to larger models. 
This negative result shows the limitation of explicit alignment objective with larger-scale bitext and encoders.

\section{Explicit Alignment Objectives}
We begin with a presentation of objective functions that use parallel data across languages for training multilingual encoders.
These objectives assume multilingual data in the form of word pairs in parallel sentences. Since gold word alignments are scarce, we use an unsupervised word aligner. Let \src and \trg be the contextual hidden state matrix of corresponding words from a pretrained multilingual encoder. We assume \src is English while \trg is a combination of different target languages. As both mBERT and XLMR operate at the subword level, we use the representation of the first subword, which is consistent with the evaluation stage. Each $s_i$ and $t_i$ are a corresponding row of \src and $\mathbf{T}$, respectively. \src and \trg come from the final layer of the encoder while $\mathbf{S}^{l}$ and $\mathbf{T}^{l}$ come from the $l^{\text{th}}$-layer.

\paragraph{Linear Mapping} If \src and \trg are static feature (such as from ELMo \citep{peters-etal-2018-deep}) then \trg can be aligned so that it is close to \src via a linear mapping \citep{wang-etal-2019-cross,wang2019cross,wu2019emerging,liu-etal-2019-investigating}, similar to aligning monolingual embeddings to produce cross-lingual embeddings. For feature $\mathbf{S}^{l}$ and $\mathbf{T}^{l}$ from layer $l$, we can learn a mapping $\mathbf{W}^{l}$.
\begin{equation}\label{eq:linear}
\mathbf{W}^{l*} = \arg\min_{\mathbf{W}^{l}} \| \mathbf{S}^{l} - \mathbf{T}^{l}\mathbf{W}^{l} \|^2_2 
\end{equation}
When $\mathbf{W}^{l}$ is orthogonal, \cref{eq:linear} is known as Procrustes problem \citep{smith2017offline} and can be solved by SVD. Alternatively, \cref{eq:linear} can also be solved by gradient descent, without the need to store in memory huge matrices \src and $\mathbf{T}$. We adopt the latter more memory efficient approach. Following \citet{conneau2017word}, we enforce the orthogonality by alternating the gradient update and the following update rule
\begin{equation}
\mathbf{W} \leftarrow (1+\beta) \mathbf{W} - \beta (\mathbf{W}\mathbf{W}^T)\mathbf{W}
\end{equation}
with $\beta = 0.01$. Note we learn different $\mathbf{W}^{l}$ for each target language.

This approach has yielded improvements in several studies.
\citet{wang-etal-2019-cross} used mBERT and 10k parallel sentences from Europarl to improve dependency parsing.
\citet{wang2019cross} used mBERT and 30k parallel sentences from Europarl to improve named entity recognition (NER) on Spanish, Dutch, and German. \citet{wu2019emerging} used bilingual BERT and 10k parallel sentences from XNLI \citep{conneau-etal-2018-xnli} to improve dependency parsing (but not NER) on French, Russian, and Chinese. \citet{liu-etal-2019-investigating} did not evaluate on cross-lingual transfer tasks.

\paragraph{L2 Alignment} Instead of using \src and \trg as static features, \citet{cao2020multilingual} proposed fine-tuning the entire encoder
\begin{equation}\label{eq:l2}
\mathcal{L}_\text{L2} (\theta) = \text{mean}_i( \| s_i - t_i \|^2_2 )
\end{equation}
where $\theta$ is the encoder parameters. 
To prevent a degenerative solution, they additionally use a regularization term
\begin{equation}\label{eq:src-hidden}
\mathcal{L}_\text{reg-hidden} (\theta) = \| \bar{\mathbf{S}} - \bar{\mathbf{S}}_\text{pretrained} \|^2_2
\end{equation}
where $\bar{\mathbf{S}}$ denote \textbf{all} hidden states of the source sentence including unaligned words, encouraging the source hidden states to stay close to the pretrained hidden states. With mBERT and 20k to 250k parallel sentences from Europarl and MultiUN, \citeauthor{cao2020multilingual} show improvement on XNLI but not parsing.\footnote{The authors state they did not observe improvements on parsing in the NLP Hightlights podcast (\#112) \citep{ai2_2020}.}

In preliminary experiments, we found constraining parameters to stay close to their original pretrained values also prevents degenerative solutions
\begin{equation}\label{eq:src-param}
\mathcal{L}_\text{reg-param} (\theta) = \| \theta - \theta_\text{pretrained} \|^2_2
\end{equation}
while being more efficient than \cref{eq:src-hidden}. As a result, we adopt the following objective (with $\lambda=1$):
\begin{equation}\label{eq:l2_and_reg}
\mathcal{L} (\theta) = \mathcal{L}_\text{L2} (\theta) + \lambda \mathcal{L}_\text{reg-param} (\theta)
\end{equation}

\subsection{Contrastive Alignment}
Inspired by the contrastive learning framework of \citet{chen2020simple}, we propose a contrastive loss to align \src and \trg by fine-tuning the encoder. Assume in each batch, we have corresponding $(s_i, t_i)$ where $i \in \{1,\dots,B\}$.
Instead of optimizing the absolute distance between $s_i$ and $t_i$ like \cref{eq:linear} or \cref{eq:l2}, contrastive loss allows more flexibility by encouraging $s_i$ and $t_i$ to be closer as compared with any other hidden state. In other words, our proposed contrastive alignment optimizes the relative distance between $s_i$ and $t_i$. As the alignment signal is often suboptimal, our alignment objective is more robust to errors in unsupervised word-level alignment. Additionally, unlike previous works, we select different sets of negative examples to enforce different levels of cross-lingual alignment. Finally, it naturally scales to multiple languages.

\paragraph{Weak alignment} When the negative examples only come from target languages, we enforce a weak cross-lingual alignment, i.e. $s_i$ should be closer to $t_i$ than any other $t_j, \forall j\neq i$. The same is true in the other direction. The loss of a batch is
\begin{align}
\mathcal{L}_\text{weak} (\theta) &\nonumber \\
= \frac{1}{2B} \sum^{B}_{i=1} ( &\log\frac{ \exp (\text{sim}(s_i, t_i ) / T) } { \sum^{B}_{j=1} \exp (\text{sim}(s_i, t_j ) / T) } \nonumber \\
+ &\log\frac{ \exp (\text{sim}(s_i, t_i ) / T) } { \sum^{B}_{j=1} \exp (\text{sim}(s_j, t_i ) / T) } )
\end{align}
where $T = 0.1$ is a temperature hyperparameter and $\text{sim}(a, b)$ measures the similarity of $a$ and $b$.

We use a learned cosine similarity $\text{sim}(a, b) = \cos(f(a), f(b))$ where $f$ is a feed-forward feature extractor with one hidden layer (768-768-128) and ReLU. It can learn to discard language-specific information and only align the align-able information. \citet{chen2020simple} find that this similarity measure learns better representation for computer vision. After alignment, $f$ is discarded as most cross-lingual transfer tasks do not need this feature extractor, though tasks like parallel sentence retrieval might find it helpful. This learned similarity cannot be applied to an absolute distance objective like \cref{eq:l2} as it can produce degenerate solutions.

\paragraph{Strong alignment} If the negative examples include both source and target languages, we enforce a strong cross-lingual alignment, i.e. $s_i$ should be closer to $t_i$ than any other $t_j, \forall j \neq i$ and $s_j, \forall j \neq i$.
\begin{align}
&\mathcal{L}_\text{strong} (\theta) \nonumber \\
= &\frac{1}{2B} \sum_{h\in \mathcal{H}} \log\frac{ \exp (\text{sim}(h, \text{aligned}(h) ) / T) } { \sum_{h'\in \mathcal{H}, h'\neq h} \exp (\text{sim}(h, h') / T) }
\end{align}
where $\text{aligned}(h)$ is the aligned hidden state of $h$ and $\mathcal{H} = \{s_1, \dots, s_B, t_1, \dots, t_B\}$.

For both weak and strong alignment objectives, we add a regularization term \cref{eq:src-param} with $\lambda = 1$.

\section{Experiments}

\insertAverTable

\paragraph{Multilingual Alignment} We consider alignment and transfer from English to 8 target languages: Arabic, German, English, Spanish, French, Hindi, Russian, Vietnamese, and Chinese. We use two sets of bitexts: (1) bitext used in previous works \citep{lample2019cross} and (2) the OPUS-100 bitext \citep{zhang2020improving}. (1) For bitext used in previous works, we use MultiUN for Arabic, Spanish, French, Russian or Chinese, EUBookshop \citep{skadins-etal-2014-billions} for German, IIT Bombay corpus \citep{kunchukuttan-etal-2018-iit} for Hindi and OpenSubtitles \citep{lison-etal-2018-opensubtitles2018} for Vietnamese. We sample 1M bitext for each target language. (2) The OPUS-100 covering 100 languages with English as the center, and sampled from the OPUS collection randomly, which better reflects the average quality of bitext for most languages. It contains 1M bitext for each target language, except Hindi (0.5M).

We tokenize the bitext with Moses \citep{koehn-etal-2007-moses} and segment Chinese with \citet{chang-etal-2008-optimizing}. %
We use \texttt{fast\_align} \citep{dyer-etal-2013-simple} to produce unsupervised word alignments in both direction and symmetrize with the \textit{grow-diag-final-and} heuristic. We only keep one-to-one alignment and discard any trivial alignment where the source and target words are identical. %

We train the L2, weak, and strong alignment objectives in a multilingual fashion. Each batch contains examples from all target languages. Following \citet{devlin-etal-2019-bert}, we optimize with Adam \citep{kingma2014adam}, learning rate $\texttt{1e-4}$, 128 batch size, 100k total steps ($\approx$ 2 epochs), 4k steps linear warmup and linear decay. We use 16-bit precision and train each model on a single RTX TITAN for around 18 hours. We set the maximum sequence length to 96. For linear mapping, we use a linear decay learning rate from $\texttt{1e-4}$ to $0$ in 20k steps ($\approx$ 3 epochs), and train for 3 hours for each language pairs.

\paragraph{Evaluation} We consider zero-shot cross-lingual transfer with XNLI \citep{conneau-etal-2018-xnli}, NER \citep{pan-etal-2017-cross}, POS tagging and dependency parsing \citep{11234/1-3226}.\footnote{We use the following treebanks: Arabic-PADT, German-GSD, English-EWT, Spanish-GSD, French-GSD, Hindi-HDTB, Russian-GSD, Vietnamese-VTB, and Chinese-GSD.} We evaluate XNLI and POS tagging with accuracy (ACC), NER with span-level F1, and parsing with labeled attachment score (LAS). For the task-specific layer, we use a linear classifier for XNLI, NER, and POS tagging, and use \citet{dozat2016deep} for dependency parsing. We fine-tune all parameters on English training data and directly transfer to target languages. We optimize with Adam, learning rate $\texttt{2e-5}$ with 10\% steps linear warmup and linear decay, 5 epochs, and 32 batch size. For the linear mapping alignment, we use an ELMo-style feature-based model\footnote{We take the weighted average of representations in all layers of the encoder.} with 4 extra Transformer layers \citep{vaswani2017attention}, a CRF instead of a linear classifier for NER, and train for 20 epochs, a batch size of 128 and learning rate $\texttt{1e-3}$ (except NER and XNLI with $\texttt{1e-4}$). All token level tasks use the first subword as the word representation for task-specific layers following previous work \citep{devlin-etal-2019-bert,wu-dredze-2019-beto}. Model selection is done on the English dev set. We report the mean and standard derivation of test performance of 5 evaluation runs with different random seeds\footnote{We pick 5 random seeds before the experiment and use the same seeds for each task and model.} and the same hyperparameters. Additional experiments detail can be found in \cref{sec:appendix}.

\section{Result}

\paragraph{Robustness of Previous Methods} With a more robust evaluation scheme and 1 million parallel sentences (4$\times$ to 100$\times$ of previously considered data), the previously proposed Linear Mapping or L2 Alignment does not consistently outperform a no alignment setting more than one standard derivation in all cases (\cref{tab:aver}).
With mBERT, L2 Alignment performs comparably to no alignment on all 4 tasks (XNLI, NER, POS tagging, and parsing). Compared to no alignment, Linear Mapping performs much worse on NER, performs better on POS tagging and parsing, and performs comparably on XNLI. While previous work observes small improvements on selected languages and tasks, it likely depends on the randomness during evaluation. Based on a more comprehensive evaluation including 4 tasks and multiple seeds, the previously proposed methods do not consistently perform better than no alignment with millions of parallel sentences.

\paragraph{Contrastive Alignment} In \cref{tab:aver}, with mBERT, both proposed contrastive alignment methods consistently perform as well as no alignment while outperforming more than 1 standard derivation on POS tagging and/or parsing. This suggests the proposed methods are more robust to suboptimal alignments. We hypothesize that learned cosine similarity and contrastive alignment allow the model to recover from suboptimal alignments. Both weak and strong alignment perform comparably. While preliminary experiments found that increasing the batch size by 1.5$\times$ does not lead to better performance, future work could consider using a memory bank to greatly increase the number of negative examples \citep{chen2020improved}, which has been shown to be beneficial for computer vision tasks.

\paragraph{Alignment with XLMR} XLMR, trained on 2.5TB of text, has the same number of transformer layers
as mBERT but larger vocabulary. It performs much better than mBERT. Therefore, we wonder if an explicit alignment objective can similarly lead to better cross-lingual representations. Unfortunately, in \cref{tab:aver}, we find all alignment methods we consider do not improve over no alignment. Compared to no alignment, Linear Mapping and L2 Alignment have worse performance in 3 out of 4 tasks (except POS tagging). In contrast to previous work, both contrastive alignment objectives perform comparably to no alignment in all 4 tasks.

\paragraph{Impact of Bitext Quality} Even though the OPUS-100 bitext has lower quality compared to bitext used in previous works (due to its greater inclusion of bitext from various sources), it has minimum impact on each alignment method we consider. This is good news for the lower resource languages, as not all languages are covered by MultiUN or Europarl.

\paragraph{Model Capacity vs Alignment} XLMR$_\text{large}$ has nearly twice the number of parameters as XLMR$_\text{base}$. Even trained on the same data, it performs much better than XLMR$_\text{base}$, with or without alignment. This suggests increasing model capacity likely leads to better cross-lingual representations than using an explicit alignment objective. Future work could tackle the curse of multilinguality \citep{conneau2019unsupervised} by increasing the model capacity in a computationally efficient way \citep{pfeiffer2020mad}.

\section{Discussion}

Our proposed contrastive alignment objective outperforms L2 Alignment \citep{cao2020multilingual} and consistently performs as well as or better than no alignment
using various quality bitext 
on 4 NLP tasks 
under a comprehensive evaluation with multiple seeds.
However, to our surprise, previously proposed methods do not show consistent improvement over no alignment in this setting.
Therefore, we make the following recommendations for future work on cross-lingual alignment or multilingual representations: 1) Evaluations should consider average quality data, not exclusively high-quality bitext. 2) Evaluation must consider multiple NLP tasks or datasets. 3) Evaluation should report \textbf{mean and variance over multiple seeds}, not a single run. More broadly, the community must establish a robust evaluation scheme for zero-shot cross-lingual transfer as a single run with one random seed does not reflect the variance of the method (especially in a zero-shot or few-shot setting).\footnote{This includes recently compiled zero-shot cross-lingual transfer benchmarks like XGLUE \citep{liang2020xglue} and XTREME \citep{hu2020xtreme}.} While \citet{keung2020evaluation} advocate using oracle for model selection, we instead argue reporting the variance of test performance, following the few-shot learning literature. Additionally, no alignment methods improve XLMR and larger XLMR$_\text{large}$ performs much better, and raw text is easier to obtain than bitext. Therefore, scaling models to more raw text and larger capacity models may be more beneficial for producing better cross-lingual models.

\section*{Acknowledgments}
This research is supported in part by ODNI, IARPA, via the BETTER Program contract \#2019-19051600005. The views and conclusions contained herein are those of the authors and should not be interpreted as necessarily representing the official policies, either expressed or implied, of ODNI, IARPA, or the U.S. Government. The U.S. Government is authorized to reproduce and distribute reprints for governmental purposes notwithstanding any copyright annotation therein.

This research is supported by the following open-source softwares: NumPy \citep{2020NumPy-Array}, PyTorch \citep{paszke2017automatic}, PyTorch lightning \cite{falcon2019pytorch}, scikit-learn \citep{JMLR:v12:pedregosa11a}, Transformer \citep{Wolf2019HuggingFacesTS}.

\bibliography{anthology,emnlp2020}
\bibliographystyle{acl_natbib}

\clearpage

\appendix

\section{Additional Experiments Detail}
\label{sec:appendix}

\paragraph{Evaluation Detail}
We set the maximum sequence length to 128 during fine-tuning. For NER and POS tagging, we additionally use a sliding window of context to include subwords beyond the first 128. At test time, we use the same maximum sequence length except for parsing. At test time for parsing, we only use the first 128 words of a sentence instead of subwords to make sure we compare different models consistently. We ignore words with POS tags of \texttt{SYM} and \texttt{PUNCT} during parsing evaluation. We rewrite the \texttt{BIO} label, similar to an unbiased structure predictor, to make sure a valid span is produced during NER evaluation. As the supervision on Chinese NER is on character-level, we segment the character into word using the Stanford Word Segmenter and realign the label.

All datasets we used are publicly available: NER\footnote{\url{https://www.amazon.com/clouddrive/share/d3KGCRCIYwhKJF0H3eWA26hjg2ZCRhjpEQtDL70FSBN}}, XNLI\footnote{\url{https://cims.nyu.edu/~sbowman/multinli/multinli_1.0.zip}}\footnote{\url{https://dl.fbaipublicfiles.com/XNLI/XNLI-1.0.zip}}, POS tagging and dependency parsing\footnote{\url{https://lindat.mff.cuni.cz/repository/xmlui/handle/11234/1-3226}}.
Data statistic can be found in \cref{tab:stat}.

\insertStatTable

\section{Breakdown of Zero-shot Cross-lingual Transfer Result}
\label{sec:breakdown}

Breakdown of alignment with bitext from previous works can be found in \cref{tab:all-xlm} and breakdown of alignment with the OPUS-100 bitext can be found in \cref{tab:all-opus}.

\insertAllXLMTable

\insertAllOpusTable

\end{document}

%% file: table.tex
\newcommand{\insertAllXLMTable}{
\begin{table*}[h]
\begin{center}
\resizebox{1\linewidth}{!}{
\begin{tabular}[b]{l|ccc ccc ccc|c}
\toprule

 & \textbf{ar} & \textbf{de} & \textbf{en} & \textbf{es} & \textbf{fr} & \textbf{hi} & \textbf{ru} & \textbf{vi} & \textbf{zh} & \textbf{AVER} \\

\midrule
\multicolumn{11}{l}{\textbf{XNLI (Accuracy)}} \\
\midrule
mBERT & 64.2$_{\pm 0.9}$ & 70.5$_{\pm 0.2}$ & 82.5$_{\pm 0.3}$ & 74.2$_{\pm 1.2}$ & 73.8$_{\pm 0.8}$ & 59.4$_{\pm 0.7}$ & 68.3$_{\pm 0.9}$ & 69.6$_{\pm 0.7}$ & 68.6$_{\pm 0.9}$ & 70.1$_{\pm 0.8}$ \\
+ Linear Mapping & 63.8$_{\pm 0.6}$ & 70.4$_{\pm 0.4}$ & \worse 81.0$_{\pm 0.5}$ & 73.9$_{\pm 0.9}$ & \worse 72.5$_{\pm 0.8}$ & \better 61.2$_{\pm 0.7}$ & \worse 67.1$_{\pm 0.4}$ & 70.2$_{\pm 0.5}$ & \better 70.1$_{\pm 0.8}$ & 70.0$_{\pm 0.6}$ \\
+ L2 Align & 64.1$_{\pm 0.4}$ & \worse 70.0$_{\pm 0.7}$ & 82.2$_{\pm 0.4}$ & 73.9$_{\pm 0.5}$ & 73.8$_{\pm 0.2}$ & \worse 58.5$_{\pm 0.3}$ & 67.9$_{\pm 0.4}$ & 69.4$_{\pm 0.6}$ & 67.9$_{\pm 0.4}$ & 69.7$_{\pm 0.4}$ \\
+ Weak Align (Our) & 64.9$_{\pm 0.8}$ & \better 71.0$_{\pm 0.8}$ & 82.3$_{\pm 0.4}$ & 74.6$_{\pm 0.7}$ & 73.8$_{\pm 0.4}$ & 59.8$_{\pm 0.3}$ & 68.5$_{\pm 1.0}$ & 70.3$_{\pm 0.8}$ & 69.4$_{\pm 1.0}$ & 70.5$_{\pm 0.7}$ \\
+ Strong Align (Our) & 64.8$_{\pm 0.8}$ & 70.5$_{\pm 0.9}$ & 82.3$_{\pm 0.5}$ & 74.4$_{\pm 0.6}$ & 74.1$_{\pm 0.7}$ & 59.8$_{\pm 0.9}$ & 68.2$_{\pm 0.6}$ & 70.1$_{\pm 0.8}$ & 69.0$_{\pm 1.0}$ & 70.4$_{\pm 0.7}$ \\

\midrule
XLMR$_\text{base}$ & 71.8$_{\pm 0.2}$ & 77.3$_{\pm 0.5}$ & 85.1$_{\pm 0.3}$ & 79.3$_{\pm 0.5}$ & 78.8$_{\pm 0.4}$ & 70.3$_{\pm 0.6}$ & 75.9$_{\pm 0.5}$ & 74.8$_{\pm 0.4}$ & 74.1$_{\pm 0.5}$ & 76.4$_{\pm 0.5}$ \\
+ Linear Mapping & \worse 69.7$_{\pm 0.6}$ & \worse 74.3$_{\pm 0.3}$ & \worse 82.5$_{\pm 0.6}$ & \worse 76.4$_{\pm 0.5}$ & \worse 75.5$_{\pm 0.4}$ & \worse 67.2$_{\pm 0.9}$ & \worse 73.2$_{\pm 0.3}$ & \worse 72.5$_{\pm 0.5}$ & \worse 68.9$_{\pm 1.2}$ & \worse 73.4$_{\pm 0.6}$ \\
+ L2 Align & 71.6$_{\pm 0.8}$ & \worse 76.0$_{\pm 0.5}$ & \worse 84.5$_{\pm 0.5}$ & \worse 78.6$_{\pm 0.3}$ & \worse 77.9$_{\pm 0.3}$ & 69.8$_{\pm 0.7}$ & \worse 75.3$_{\pm 0.3}$ & \worse 74.0$_{\pm 0.4}$ & 73.7$_{\pm 0.7}$ & \worse 75.7$_{\pm 0.5}$ \\
+ Weak Align (Our) & 71.7$_{\pm 0.7}$ & \worse 76.5$_{\pm 0.6}$ & \worse 84.7$_{\pm 0.6}$ & \worse 78.7$_{\pm 0.6}$ & \worse 78.1$_{\pm 0.7}$ & 70.4$_{\pm 0.9}$ & 75.8$_{\pm 0.6}$ & 74.5$_{\pm 0.5}$ & 74.2$_{\pm 0.7}$ & 76.1$_{\pm 0.7}$ \\
+ Strong Align (Our) & 71.6$_{\pm 0.5}$ & \worse 76.6$_{\pm 0.4}$ & \worse 84.7$_{\pm 0.5}$ & 79.0$_{\pm 0.4}$ & \worse 78.3$_{\pm 0.3}$ & 70.0$_{\pm 1.0}$ & 75.7$_{\pm 0.7}$ & 74.7$_{\pm 0.4}$ & 73.7$_{\pm 0.8}$ & 76.0$_{\pm 0.6}$ \\

\midrule
XLMR$_{\text{large}}$ & 77.5$_{\pm 0.6}$ & 81.7$_{\pm 0.4}$ & 88.0$_{\pm 0.3}$ & 83.3$_{\pm 0.6}$ & 82.0$_{\pm 0.5}$ & 75.1$_{\pm 0.8}$ & 79.2$_{\pm 0.7}$ & 78.4$_{\pm 0.6}$ & 78.3$_{\pm 0.6}$ & 80.4$_{\pm 0.6}$ \\

\midrule
\multicolumn{11}{l}{\textbf{NER (Entity-level F1)}} \\
\midrule
mBERT & 42.0$_{\pm 2.9}$ & 79.0$_{\pm 0.3}$ & 84.1$_{\pm 0.2}$ & 73.3$_{\pm 2.5}$ & 78.9$_{\pm 0.3}$ & 65.7$_{\pm 1.4}$ & 65.2$_{\pm 1.4}$ & 69.7$_{\pm 1.8}$ & 51.7$_{\pm 0.8}$ & 67.7$_{\pm 1.3}$ \\
+ Linear Mapping & \worse 36.9$_{\pm 1.1}$ & \worse 76.1$_{\pm 0.4}$ & \worse 82.8$_{\pm 0.1}$ & \worse 70.4$_{\pm 2.1}$ & \worse 77.4$_{\pm 0.7}$ & 64.5$_{\pm 1.4}$ & \worse 59.5$_{\pm 2.5}$ & \worse 65.2$_{\pm 2.7}$ & \worse 40.5$_{\pm 2.0}$ & \worse 63.7$_{\pm 1.5}$ \\
+ L2 Align & 39.7$_{\pm 1.6}$ & \worse 77.7$_{\pm 0.8}$ & 84.0$_{\pm 0.1}$ & 72.5$_{\pm 1.5}$ & 79.1$_{\pm 0.3}$ & \worse 63.3$_{\pm 1.8}$ & 64.3$_{\pm 1.0}$ & 71.2$_{\pm 0.9}$ & 52.1$_{\pm 1.1}$ & 67.1$_{\pm 1.0}$ \\
+ Weak Align (Our) & 42.3$_{\pm 2.7}$ & 78.7$_{\pm 0.3}$ & 84.2$_{\pm 0.2}$ & 71.6$_{\pm 2.2}$ & \better 79.4$_{\pm 0.6}$ & \better 67.6$_{\pm 1.3}$ & 64.8$_{\pm 0.8}$ & 70.0$_{\pm 2.3}$ & \better 52.9$_{\pm 0.9}$ & 68.0$_{\pm 1.3}$ \\
+ Strong Align (Our) & 40.6$_{\pm 1.0}$ & \worse 78.7$_{\pm 0.3}$ & 84.2$_{\pm 0.2}$ & 72.2$_{\pm 2.5}$ & 79.0$_{\pm 0.5}$ & \better 67.2$_{\pm 0.7}$ & 64.5$_{\pm 1.7}$ & 70.1$_{\pm 2.5}$ & \better 52.5$_{\pm 0.8}$ & 67.7$_{\pm 1.1}$ \\

\midrule
XLMR$_\text{base}$ & 44.0$_{\pm 1.3}$ & 75.0$_{\pm 0.3}$ & 82.2$_{\pm 0.2}$ & 76.0$_{\pm 2.4}$ & 77.6$_{\pm 0.7}$ & 65.7$_{\pm 0.6}$ & 64.1$_{\pm 0.7}$ & 68.0$_{\pm 1.2}$ & 45.1$_{\pm 0.8}$ & 66.4$_{\pm 0.9}$ \\
+ Linear Mapping & \worse 30.8$_{\pm 2.1}$ & \worse 69.0$_{\pm 0.6}$ & \worse 78.3$_{\pm 0.3}$ & \worse 59.8$_{\pm 0.5}$ & \worse 67.8$_{\pm 0.7}$ & \worse 57.9$_{\pm 1.5}$ & \worse 48.0$_{\pm 1.0}$ & \worse 54.4$_{\pm 0.5}$ & \worse 21.0$_{\pm 0.9}$ & \worse 54.1$_{\pm 0.9}$ \\
+ L2 Align & 44.9$_{\pm 2.1}$ & 74.9$_{\pm 0.6}$ & 82.1$_{\pm 0.3}$ & 75.0$_{\pm 3.1}$ & 77.1$_{\pm 0.6}$ & 65.5$_{\pm 1.3}$ & \worse 63.2$_{\pm 0.3}$ & \worse 66.3$_{\pm 2.2}$ & \worse 42.4$_{\pm 0.7}$ & 65.7$_{\pm 1.2}$ \\
+ Weak Align (Our) & \better 45.6$_{\pm 1.4}$ & 75.0$_{\pm 0.5}$ & 82.2$_{\pm 0.2}$ & 74.2$_{\pm 2.4}$ & 77.2$_{\pm 0.8}$ & 65.8$_{\pm 1.1}$ & 63.6$_{\pm 1.1}$ & 67.6$_{\pm 0.7}$ & \worse 42.8$_{\pm 0.6}$ & 66.0$_{\pm 1.0}$ \\
+ Strong Align (Our) & \better 45.7$_{\pm 1.7}$ & 75.1$_{\pm 0.6}$ & 82.1$_{\pm 0.3}$ & \worse 73.5$_{\pm 1.7}$ & 77.2$_{\pm 0.6}$ & 65.8$_{\pm 1.7}$ & 63.7$_{\pm 0.5}$ & 68.1$_{\pm 0.8}$ & \worse 43.2$_{\pm 0.4}$ & 66.1$_{\pm 0.9}$ \\

\midrule
XLMR$_{\text{large}}$ & 46.8$_{\pm 4.3}$ & 79.1$_{\pm 0.5}$ & 84.2$_{\pm 0.2}$ & 75.7$_{\pm 2.9}$ & 80.7$_{\pm 0.5}$ & 71.6$_{\pm 1.1}$ & 71.7$_{\pm 0.5}$ & 77.4$_{\pm 1.3}$ & 51.5$_{\pm 1.4}$ & 71.0$_{\pm 1.4}$ \\

\midrule
\multicolumn{11}{l}{\textbf{POS (Accuracy)}} \\
\midrule
mBERT & 60.3$_{\pm 0.9}$ & 90.4$_{\pm 0.3}$ & 96.9$_{\pm 0.1}$ & 87.7$_{\pm 0.2}$ & 88.9$_{\pm 0.3}$ & 68.0$_{\pm 0.8}$ & 82.5$_{\pm 0.7}$ & 62.7$_{\pm 0.2}$ & 67.1$_{\pm 1.1}$ & 78.3$_{\pm 0.5}$ \\
+ Linear Mapping & \better 73.6$_{\pm 0.7}$ & \worse 88.2$_{\pm 0.5}$ & \worse 96.3$_{\pm 0.0}$ & \worse 87.4$_{\pm 0.1}$ & 88.9$_{\pm 0.3}$ & \better 77.3$_{\pm 0.6}$ & \worse 78.0$_{\pm 1.0}$ & \worse 60.4$_{\pm 0.5}$ & \worse 65.7$_{\pm 1.3}$ & \better 79.5$_{\pm 0.5}$ \\
+ L2 Align & \better 63.4$_{\pm 2.6}$ & \worse 89.3$_{\pm 0.7}$ & \worse 96.7$_{\pm 0.2}$ & \worse 86.7$_{\pm 0.3}$ & \worse 87.9$_{\pm 0.5}$ & \worse 65.2$_{\pm 3.9}$ & \better 83.6$_{\pm 0.9}$ & \worse 62.3$_{\pm 0.8}$ & 66.5$_{\pm 1.5}$ & 78.0$_{\pm 1.3}$ \\
+ Weak Align (Our) & \better 61.6$_{\pm 2.0}$ & 90.3$_{\pm 0.7}$ & 96.9$_{\pm 0.1}$ & 87.5$_{\pm 0.6}$ & \worse 88.6$_{\pm 0.3}$ & \better 70.3$_{\pm 0.9}$ & 83.1$_{\pm 0.6}$ & \better 63.2$_{\pm 0.3}$ & 68.1$_{\pm 0.9}$ & \better 78.8$_{\pm 0.7}$ \\
+ Strong Align (Our) & \better 61.9$_{\pm 2.0}$ & 90.4$_{\pm 0.7}$ & 96.9$_{\pm 0.0}$ & 87.5$_{\pm 0.5}$ & \worse 88.5$_{\pm 0.4}$ & \better 71.1$_{\pm 1.2}$ & 83.0$_{\pm 0.5}$ & \better 63.2$_{\pm 0.2}$ & 68.0$_{\pm 0.6}$ & \better 79.0$_{\pm 0.7}$ \\

\midrule
XLMR$_\text{base}$ & 70.2$_{\pm 1.6}$ & 91.6$_{\pm 0.3}$ & 97.5$_{\pm 0.0}$ & 88.5$_{\pm 0.2}$ & 89.4$_{\pm 0.3}$ & 71.7$_{\pm 1.3}$ & 86.1$_{\pm 0.3}$ & 64.5$_{\pm 0.5}$ & 71.4$_{\pm 0.5}$ & 81.2$_{\pm 0.6}$ \\
+ Linear Mapping & \better 74.3$_{\pm 1.1}$ & \worse 90.7$_{\pm 0.5}$ & \worse 96.9$_{\pm 0.0}$ & \worse 88.2$_{\pm 0.1}$ & 89.3$_{\pm 0.3}$ & \better 82.1$_{\pm 0.9}$ & \worse 82.7$_{\pm 0.4}$ & \worse 62.6$_{\pm 0.4}$ & \worse 65.3$_{\pm 1.0}$ & 81.3$_{\pm 0.5}$ \\
+ L2 Align & 71.1$_{\pm 1.8}$ & 91.4$_{\pm 0.3}$ & \worse 97.4$_{\pm 0.0}$ & \worse 88.2$_{\pm 0.2}$ & \worse 89.0$_{\pm 0.3}$ & 73.0$_{\pm 3.8}$ & \better 86.6$_{\pm 0.2}$ & 64.4$_{\pm 0.4}$ & \worse 70.8$_{\pm 0.8}$ & 81.3$_{\pm 0.9}$ \\
+ Weak Align (Our) & \better 72.8$_{\pm 0.7}$ & \worse 91.1$_{\pm 0.2}$ & \worse 97.4$_{\pm 0.0}$ & 88.3$_{\pm 0.2}$ & 89.2$_{\pm 0.2}$ & 72.4$_{\pm 1.6}$ & 86.4$_{\pm 0.1}$ & 64.7$_{\pm 0.4}$ & 71.6$_{\pm 1.2}$ & 81.5$_{\pm 0.5}$ \\
+ Strong Align (Our) & \better 72.5$_{\pm 0.9}$ & \worse 91.1$_{\pm 0.3}$ & \worse 97.4$_{\pm 0.0}$ & \worse 88.3$_{\pm 0.2}$ & \worse 89.1$_{\pm 0.1}$ & 72.0$_{\pm 2.1}$ & 86.4$_{\pm 0.1}$ & 64.8$_{\pm 0.4}$ & 71.4$_{\pm 1.1}$ & 81.4$_{\pm 0.6}$ \\

\midrule
XLMR$_{\text{large}}$ & 73.9$_{\pm 1.0}$ & 91.9$_{\pm 0.3}$ & 98.0$_{\pm 0.0}$ & 89.2$_{\pm 0.2}$ & 89.8$_{\pm 0.1}$ & 78.4$_{\pm 2.1}$ & 86.5$_{\pm 0.2}$ & 64.8$_{\pm 0.3}$ & 71.0$_{\pm 0.3}$ & 82.6$_{\pm 0.5}$ \\

\midrule
\multicolumn{11}{l}{\textbf{Parsing (Labeled Attachment Score)}} \\
\midrule
mBERT & 28.8$_{\pm 0.4}$ & 67.8$_{\pm 0.5}$ & 79.7$_{\pm 0.1}$ & 69.1$_{\pm 0.1}$ & 73.3$_{\pm 0.2}$ & 31.0$_{\pm 0.5}$ & 60.2$_{\pm 0.6}$ & 33.5$_{\pm 0.5}$ & 29.5$_{\pm 0.4}$ & 52.6$_{\pm 0.4}$ \\
+ Linear Mapping & \better 44.1$_{\pm 0.3}$ & \worse 64.4$_{\pm 0.4}$ & \better 80.5$_{\pm 0.2}$ & \better 70.2$_{\pm 0.3}$ & \better 73.9$_{\pm 0.1}$ & \better 32.2$_{\pm 0.3}$ & \worse 56.7$_{\pm 0.5}$ & \worse 32.1$_{\pm 0.2}$ & \worse 28.1$_{\pm 0.3}$ & \better 53.6$_{\pm 0.3}$ \\
+ L2 Align & \better 29.6$_{\pm 1.6}$ & \worse 66.9$_{\pm 0.2}$ & \worse 79.2$_{\pm 0.2}$ & \worse 68.2$_{\pm 0.4}$ & \worse 72.5$_{\pm 0.5}$ & 30.8$_{\pm 1.9}$ & 60.0$_{\pm 0.6}$ & 33.3$_{\pm 0.4}$ & 29.5$_{\pm 0.4}$ & 52.2$_{\pm 0.7}$ \\
+ Weak Align (Our) & \better 30.7$_{\pm 0.9}$ & 67.6$_{\pm 0.6}$ & \better 79.8$_{\pm 0.1}$ & \better 69.7$_{\pm 0.4}$ & \better 73.6$_{\pm 0.4}$ & 31.2$_{\pm 0.8}$ & \better 61.3$_{\pm 0.7}$ & 33.5$_{\pm 0.6}$ & \better 30.5$_{\pm 0.6}$ & \better 53.1$_{\pm 0.6}$ \\
+ Strong Align (Our) & \better 31.2$_{\pm 1.1}$ & 67.5$_{\pm 0.4}$ & 79.8$_{\pm 0.1}$ & \better 69.4$_{\pm 0.3}$ & 73.4$_{\pm 0.5}$ & 30.7$_{\pm 1.5}$ & \better 61.3$_{\pm 0.8}$ & 33.5$_{\pm 0.6}$ & \better 30.0$_{\pm 0.5}$ & \better 53.0$_{\pm 0.6}$ \\

\midrule
XLMR$_\text{base}$ & 43.7$_{\pm 1.7}$ & 69.0$_{\pm 0.4}$ & 80.5$_{\pm 0.2}$ & 71.0$_{\pm 0.4}$ & 73.6$_{\pm 0.5}$ & 41.2$_{\pm 0.9}$ & 66.3$_{\pm 0.9}$ & 36.6$_{\pm 0.2}$ & 34.2$_{\pm 0.7}$ & 57.3$_{\pm 0.6}$ \\
+ Linear Mapping & \better 47.2$_{\pm 0.6}$ & \worse 66.7$_{\pm 0.3}$ & \better 81.4$_{\pm 0.1}$ & \better 72.6$_{\pm 0.2}$ & \better 74.4$_{\pm 0.4}$ & 41.4$_{\pm 0.7}$ & \worse 60.8$_{\pm 0.6}$ & \worse 34.3$_{\pm 0.3}$ & \worse 21.5$_{\pm 1.1}$ & \worse 55.6$_{\pm 0.5}$ \\
+ L2 Align & \worse 41.3$_{\pm 1.8}$ & \worse 68.1$_{\pm 0.3}$ & \worse 79.7$_{\pm 0.2}$ & \worse 70.0$_{\pm 0.5}$ & \worse 73.0$_{\pm 0.5}$ & \worse 40.2$_{\pm 1.6}$ & \worse 63.7$_{\pm 0.9}$ & 36.5$_{\pm 0.5}$ & \worse 32.9$_{\pm 0.3}$ & \worse 56.2$_{\pm 0.7}$ \\
+ Weak Align (Our) & 44.6$_{\pm 1.0}$ & 68.8$_{\pm 0.4}$ & 80.4$_{\pm 0.1}$ & 71.4$_{\pm 0.2}$ & 73.9$_{\pm 0.2}$ & 41.0$_{\pm 0.6}$ & 65.7$_{\pm 0.4}$ & 36.7$_{\pm 0.4}$ & 33.8$_{\pm 0.3}$ & 57.4$_{\pm 0.4}$ \\
+ Strong Align (Our) & 44.8$_{\pm 0.9}$ & 68.9$_{\pm 0.5}$ & 80.4$_{\pm 0.1}$ & 71.3$_{\pm 0.2}$ & 73.9$_{\pm 0.1}$ & 40.7$_{\pm 0.8}$ & 66.2$_{\pm 0.4}$ & 36.7$_{\pm 0.3}$ & 34.0$_{\pm 0.8}$ & 57.4$_{\pm 0.5}$ \\

\midrule
XLMR$_{\text{large}}$ & 48.2$_{\pm 1.5}$ & 67.8$_{\pm 0.6}$ & 82.6$_{\pm 0.3}$ & 73.9$_{\pm 0.4}$ & 76.4$_{\pm 0.4}$ & 41.8$_{\pm 2.5}$ & 69.6$_{\pm 0.4}$ & 38.9$_{\pm 0.6}$ & 35.4$_{\pm 0.5}$ & 59.4$_{\pm 0.8}$ \\

\bottomrule
\end{tabular}
}
\caption{Zero-shot cross-lingual transfer result with bitext from previous works. 
\textcolor{better}{Blue} or \textcolor{worse}{orange} indicates the mean performance is one standard derivation \textcolor{better}{above} or \textcolor{worse}{below} the mean of baseline.
\label{tab:all-xlm}}
\end{center}
\end{table*}
}

\newcommand{\insertAllOpusTable}{
\begin{table*}[h]
\begin{center}
\resizebox{1\linewidth}{!}{
\begin{tabular}[b]{l|ccc ccc ccc|c}
\toprule

 & \textbf{ar} & \textbf{de} & \textbf{en} & \textbf{es} & \textbf{fr} & \textbf{hi} & \textbf{ru} & \textbf{vi} & \textbf{zh} & \textbf{AVER} \\

\midrule
\multicolumn{11}{l}{\textbf{XNLI (Accuracy)}} \\
\midrule

mBERT & 64.2$_{\pm 0.9}$ & 70.5$_{\pm 0.2}$ & 82.5$_{\pm 0.3}$ & 74.2$_{\pm 1.2}$ & 73.8$_{\pm 0.8}$ & 59.4$_{\pm 0.7}$ & 68.3$_{\pm 0.9}$ & 69.6$_{\pm 0.7}$ & 68.6$_{\pm 0.9}$ & 70.1$_{\pm 0.8}$ \\
+ Linear Mapping & 64.1$_{\pm 0.7}$ & \worse 70.0$_{\pm 0.6}$ & \worse 81.0$_{\pm 0.5}$ & 74.1$_{\pm 0.6}$ & \worse 72.9$_{\pm 0.9}$ & \better 61.8$_{\pm 0.7}$ & \worse 67.4$_{\pm 0.6}$ & 70.2$_{\pm 0.5}$ & \better 70.2$_{\pm 0.8}$ & 70.2$_{\pm 0.6}$ \\
+ L2 Align & 64.3$_{\pm 0.5}$ & 70.7$_{\pm 1.0}$ & 82.5$_{\pm 0.5}$ & 74.3$_{\pm 0.3}$ & 74.0$_{\pm 0.4}$ & 59.3$_{\pm 0.4}$ & 68.6$_{\pm 0.7}$ & 69.7$_{\pm 0.4}$ & 69.1$_{\pm 0.5}$ & 70.3$_{\pm 0.5}$ \\
+ Weak Align (Our) & 65.1$_{\pm 0.9}$ & \better 70.9$_{\pm 0.6}$ & 82.6$_{\pm 0.5}$ & 74.9$_{\pm 0.6}$ & 74.1$_{\pm 0.4}$ & \better 60.3$_{\pm 0.6}$ & 68.9$_{\pm 0.8}$ & \better 70.6$_{\pm 0.6}$ & \better 69.6$_{\pm 1.0}$ & 70.8$_{\pm 0.7}$ \\
+ Strong Align (Our) & 64.7$_{\pm 0.9}$ & \better 70.8$_{\pm 0.7}$ & 82.4$_{\pm 0.1}$ & 74.5$_{\pm 0.7}$ & 73.9$_{\pm 0.7}$ & 59.6$_{\pm 0.6}$ & 68.5$_{\pm 1.1}$ & \better 70.4$_{\pm 0.6}$ & 69.1$_{\pm 1.0}$ & 70.4$_{\pm 0.7}$ \\

\midrule
XLMR$_\text{base}$ & 71.8$_{\pm 0.2}$ & 77.3$_{\pm 0.5}$ & 85.1$_{\pm 0.3}$ & 79.3$_{\pm 0.5}$ & 78.8$_{\pm 0.4}$ & 70.3$_{\pm 0.6}$ & 75.9$_{\pm 0.5}$ & 74.8$_{\pm 0.4}$ & 74.1$_{\pm 0.5}$ & 76.4$_{\pm 0.5}$ \\
+ Linear Mapping & \worse 69.9$_{\pm 0.4}$ & \worse 74.3$_{\pm 0.3}$ & \worse 82.5$_{\pm 0.6}$ & \worse 76.4$_{\pm 0.5}$ & \worse 75.5$_{\pm 0.6}$ & \worse 67.2$_{\pm 1.0}$ & \worse 72.7$_{\pm 0.2}$ & \worse 72.7$_{\pm 0.5}$ & \worse 70.1$_{\pm 0.8}$ & \worse 73.5$_{\pm 0.5}$ \\
+ L2 Align & 71.9$_{\pm 0.6}$ & \worse 76.4$_{\pm 0.4}$ & \worse 84.6$_{\pm 0.3}$ & \worse 78.4$_{\pm 0.5}$ & \worse 77.8$_{\pm 0.3}$ & 69.9$_{\pm 0.8}$ & \worse 75.2$_{\pm 0.5}$ & \worse 74.2$_{\pm 0.5}$ & 73.7$_{\pm 0.5}$ & \worse 75.8$_{\pm 0.5}$ \\
+ Weak Align (Our) & 71.8$_{\pm 0.6}$ & \worse 76.5$_{\pm 0.5}$ & \worse 84.6$_{\pm 0.2}$ & 79.0$_{\pm 0.4}$ & 78.4$_{\pm 0.5}$ & 70.0$_{\pm 0.5}$ & 75.7$_{\pm 0.3}$ & 74.7$_{\pm 0.3}$ & \worse 73.4$_{\pm 0.6}$ & 76.0$_{\pm 0.4}$ \\
+ Strong Align (Our) & 72.0$_{\pm 0.5}$ & \worse 76.6$_{\pm 0.4}$ & 84.8$_{\pm 0.1}$ & 79.0$_{\pm 0.4}$ & 78.6$_{\pm 0.5}$ & 70.1$_{\pm 0.3}$ & 75.7$_{\pm 0.4}$ & 74.8$_{\pm 0.6}$ & 73.8$_{\pm 0.6}$ & 76.1$_{\pm 0.4}$ \\

\midrule
XLMR$_{\text{large}}$ & 77.5$_{\pm 0.6}$ & 81.7$_{\pm 0.4}$ & 88.0$_{\pm 0.3}$ & 83.3$_{\pm 0.6}$ & 82.0$_{\pm 0.5}$ & 75.1$_{\pm 0.8}$ & 79.2$_{\pm 0.7}$ & 78.4$_{\pm 0.6}$ & 78.3$_{\pm 0.6}$ & 80.4$_{\pm 0.6}$ \\

\midrule
\multicolumn{11}{l}{\textbf{NER (Entity-level F1)}} \\
\midrule

mBERT & 42.0$_{\pm 2.9}$ & 79.0$_{\pm 0.3}$ & 84.1$_{\pm 0.2}$ & 73.3$_{\pm 2.5}$ & 78.9$_{\pm 0.3}$ & 65.7$_{\pm 1.4}$ & 65.2$_{\pm 1.4}$ & 69.7$_{\pm 1.8}$ & 51.7$_{\pm 0.8}$ & 67.7$_{\pm 1.3}$ \\
+ Linear Mapping & \worse 36.9$_{\pm 0.9}$ & \worse 76.2$_{\pm 0.3}$ & \worse 82.8$_{\pm 0.1}$ & 71.2$_{\pm 1.5}$ & \worse 77.4$_{\pm 0.7}$ & \worse 62.4$_{\pm 2.2}$ & \worse 59.6$_{\pm 2.4}$ & \worse 65.4$_{\pm 2.6}$ & \worse 42.3$_{\pm 1.4}$ & \worse 63.8$_{\pm 1.3}$ \\
+ L2 Align & 41.3$_{\pm 3.2}$ & \worse 78.2$_{\pm 1.0}$ & 84.1$_{\pm 0.1}$ & 73.4$_{\pm 2.4}$ & \better 79.7$_{\pm 0.8}$ & 64.9$_{\pm 1.5}$ & 64.9$_{\pm 1.6}$ & \better 71.8$_{\pm 0.9}$ & 52.4$_{\pm 1.3}$ & 67.8$_{\pm 1.4}$ \\
+ Weak Align (Our) & 40.3$_{\pm 1.1}$ & \worse 78.7$_{\pm 0.3}$ & 84.0$_{\pm 0.1}$ & \worse 70.7$_{\pm 2.1}$ & 79.0$_{\pm 0.4}$ & \better 67.2$_{\pm 1.2}$ & 64.9$_{\pm 1.2}$ & 69.1$_{\pm 0.8}$ & 52.0$_{\pm 1.1}$ & 67.3$_{\pm 0.9}$ \\
+ Strong Align (Our) & 40.7$_{\pm 1.9}$ & \worse 78.3$_{\pm 0.3}$ & 84.2$_{\pm 0.1}$ & \worse 70.0$_{\pm 2.6}$ & 78.8$_{\pm 0.3}$ & 66.7$_{\pm 1.4}$ & 64.8$_{\pm 0.9}$ & 69.5$_{\pm 1.4}$ & 52.1$_{\pm 0.6}$ & 67.2$_{\pm 1.1}$ \\

\midrule
XLMR$_\text{base}$ & 44.0$_{\pm 1.3}$ & 75.0$_{\pm 0.3}$ & 82.2$_{\pm 0.2}$ & 76.0$_{\pm 2.4}$ & 77.6$_{\pm 0.7}$ & 65.7$_{\pm 0.6}$ & 64.1$_{\pm 0.7}$ & 68.0$_{\pm 1.2}$ & 45.1$_{\pm 0.8}$ & 66.4$_{\pm 0.9}$ \\
+ Linear Mapping & \worse 30.8$_{\pm 1.6}$ & \worse 69.3$_{\pm 0.6}$ & \worse 78.3$_{\pm 0.3}$ & \worse 60.2$_{\pm 0.8}$ & \worse 67.9$_{\pm 0.5}$ & \worse 58.2$_{\pm 0.7}$ & \worse 47.7$_{\pm 0.8}$ & \worse 54.1$_{\pm 0.3}$ & \worse 21.6$_{\pm 1.2}$ & \worse 54.2$_{\pm 0.8}$ \\
+ L2 Align & 44.1$_{\pm 1.2}$ & \worse 74.2$_{\pm 0.7}$ & \worse 81.9$_{\pm 0.3}$ & 74.9$_{\pm 3.3}$ & \worse 76.9$_{\pm 0.6}$ & \worse 64.7$_{\pm 0.5}$ & \worse 61.9$_{\pm 1.4}$ & 68.4$_{\pm 2.2}$ & \worse 42.1$_{\pm 1.1}$ & \worse 65.5$_{\pm 1.2}$ \\
+ Weak Align (Our) & \better 45.5$_{\pm 2.8}$ & 75.0$_{\pm 0.8}$ & 82.2$_{\pm 0.2}$ & \worse 73.7$_{\pm 1.8}$ & 77.3$_{\pm 0.6}$ & \better 66.6$_{\pm 1.3}$ & 64.0$_{\pm 1.2}$ & 67.5$_{\pm 1.4}$ & \worse 43.9$_{\pm 1.2}$ & 66.2$_{\pm 1.2}$ \\
+ Strong Align (Our) & 45.3$_{\pm 1.5}$ & 75.1$_{\pm 0.4}$ & 82.2$_{\pm 0.2}$ & 74.6$_{\pm 2.5}$ & 77.4$_{\pm 0.6}$ & 66.0$_{\pm 1.2}$ & 63.7$_{\pm 0.9}$ & 68.0$_{\pm 1.1}$ & \worse 43.3$_{\pm 0.4}$ & 66.2$_{\pm 1.0}$ \\

\midrule
XLMR$_{\text{large}}$ & 46.8$_{\pm 4.3}$ & 79.1$_{\pm 0.5}$ & 84.2$_{\pm 0.2}$ & 75.7$_{\pm 2.9}$ & 80.7$_{\pm 0.5}$ & 71.6$_{\pm 1.1}$ & 71.7$_{\pm 0.5}$ & 77.4$_{\pm 1.3}$ & 51.5$_{\pm 1.4}$ & 71.0$_{\pm 1.4}$ \\

\midrule
\multicolumn{11}{l}{\textbf{POS (Accuracy)}} \\
\midrule

mBERT & 60.3$_{\pm 0.9}$ & 90.4$_{\pm 0.3}$ & 96.9$_{\pm 0.1}$ & 87.7$_{\pm 0.2}$ & 88.9$_{\pm 0.3}$ & 68.0$_{\pm 0.8}$ & 82.5$_{\pm 0.7}$ & 62.7$_{\pm 0.2}$ & 67.1$_{\pm 1.1}$ & 78.3$_{\pm 0.5}$ \\
+ Linear Mapping & \better 76.2$_{\pm 0.5}$ & \better 91.2$_{\pm 0.1}$ & \worse 96.3$_{\pm 0.0}$ & 87.6$_{\pm 0.1}$ & 89.0$_{\pm 0.2}$ & \better 74.9$_{\pm 1.1}$ & \worse 80.6$_{\pm 0.3}$ & \worse 60.4$_{\pm 0.5}$ & \worse 64.8$_{\pm 1.3}$ & \better 80.1$_{\pm 0.4}$ \\
+ L2 Align & \better 62.7$_{\pm 2.9}$ & \worse 89.5$_{\pm 0.8}$ & \worse 96.8$_{\pm 0.1}$ & \worse 87.1$_{\pm 0.3}$ & \worse 88.3$_{\pm 0.2}$ & \worse 65.2$_{\pm 3.7}$ & \better 83.8$_{\pm 1.0}$ & 62.8$_{\pm 0.5}$ & 67.3$_{\pm 1.1}$ & 78.2$_{\pm 1.2}$ \\
+ Weak Align (Our) & 61.1$_{\pm 1.3}$ & 90.4$_{\pm 0.8}$ & 96.9$_{\pm 0.0}$ & 87.7$_{\pm 0.5}$ & 88.7$_{\pm 0.3}$ & \better 70.3$_{\pm 1.2}$ & 83.2$_{\pm 0.6}$ & \better 63.3$_{\pm 0.3}$ & 68.0$_{\pm 0.5}$ & \better 78.8$_{\pm 0.6}$ \\
+ Strong Align (Our) & \better 61.7$_{\pm 1.7}$ & 90.5$_{\pm 0.7}$ & 96.9$_{\pm 0.0}$ & 87.7$_{\pm 0.6}$ & 88.7$_{\pm 0.4}$ & \better 70.5$_{\pm 1.0}$ & \better 83.3$_{\pm 0.7}$ & \better 63.1$_{\pm 0.3}$ & \better 68.2$_{\pm 0.8}$ & \better 79.0$_{\pm 0.7}$ \\

\midrule
XLMR$_\text{base}$ & 70.2$_{\pm 1.6}$ & 91.6$_{\pm 0.3}$ & 97.5$_{\pm 0.0}$ & 88.5$_{\pm 0.2}$ & 89.4$_{\pm 0.3}$ & 71.7$_{\pm 1.3}$ & 86.1$_{\pm 0.3}$ & 64.5$_{\pm 0.5}$ & 71.4$_{\pm 0.5}$ & 81.2$_{\pm 0.6}$ \\
+ Linear Mapping & \better 76.0$_{\pm 0.9}$ & \better 92.0$_{\pm 0.1}$ & \worse 96.9$_{\pm 0.0}$ & 88.7$_{\pm 0.2}$ & 89.5$_{\pm 0.3}$ & \better 78.9$_{\pm 2.1}$ & \worse 83.9$_{\pm 0.3}$ & \worse 62.5$_{\pm 0.4}$ & \worse 66.5$_{\pm 1.0}$ & 81.7$_{\pm 0.6}$ \\
+ L2 Align & 71.0$_{\pm 0.9}$ & \worse 91.2$_{\pm 0.5}$ & \worse 97.3$_{\pm 0.0}$ & \worse 87.9$_{\pm 0.3}$ & \worse 88.8$_{\pm 0.4}$ & \better 74.8$_{\pm 2.9}$ & \better 86.9$_{\pm 0.8}$ & \worse 64.0$_{\pm 0.6}$ & \worse 70.6$_{\pm 0.5}$ & 81.4$_{\pm 0.8}$ \\
+ Weak Align (Our) & \better 72.5$_{\pm 0.8}$ & \worse 91.2$_{\pm 0.3}$ & \worse 97.4$_{\pm 0.0}$ & \worse 88.2$_{\pm 0.2}$ & 89.2$_{\pm 0.2}$ & 72.7$_{\pm 1.3}$ & 86.2$_{\pm 0.2}$ & 64.7$_{\pm 0.4}$ & 71.8$_{\pm 1.4}$ & 81.5$_{\pm 0.5}$ \\
+ Strong Align (Our) & \better 72.5$_{\pm 0.6}$ & \worse 91.2$_{\pm 0.2}$ & 97.4$_{\pm 0.1}$ & 88.3$_{\pm 0.2}$ & 89.2$_{\pm 0.2}$ & 72.0$_{\pm 1.9}$ & \better 86.5$_{\pm 0.2}$ & 64.8$_{\pm 0.4}$ & 71.7$_{\pm 1.7}$ & 81.5$_{\pm 0.6}$ \\

\midrule
XLMR$_{\text{large}}$ & 73.9$_{\pm 1.0}$ & 91.9$_{\pm 0.3}$ & 98.0$_{\pm 0.0}$ & 89.2$_{\pm 0.2}$ & 89.8$_{\pm 0.1}$ & 78.4$_{\pm 2.1}$ & 86.5$_{\pm 0.2}$ & 64.8$_{\pm 0.3}$ & 71.0$_{\pm 0.3}$ & 82.6$_{\pm 0.5}$ \\

\midrule
\multicolumn{11}{l}{\textbf{Parsing (Labeled Attachment Score)}} \\
\midrule

mBERT & 28.8$_{\pm 0.4}$ & 67.8$_{\pm 0.5}$ & 79.7$_{\pm 0.1}$ & 69.1$_{\pm 0.1}$ & 73.3$_{\pm 0.2}$ & 31.0$_{\pm 0.5}$ & 60.2$_{\pm 0.6}$ & 33.5$_{\pm 0.5}$ & 29.5$_{\pm 0.4}$ & 52.6$_{\pm 0.4}$ \\
+ Linear Mapping & \better 45.0$_{\pm 0.3}$ & 67.7$_{\pm 0.2}$ & \better 80.5$_{\pm 0.2}$ & \better 70.0$_{\pm 0.3}$ & \better 73.9$_{\pm 0.2}$ & \worse 28.4$_{\pm 0.2}$ & \worse 57.2$_{\pm 0.4}$ & \worse 32.0$_{\pm 0.3}$ & \worse 28.1$_{\pm 0.2}$ & \better 53.6$_{\pm 0.3}$ \\
+ L2 Align & \better 29.7$_{\pm 0.6}$ & 67.7$_{\pm 0.7}$ & \worse 79.3$_{\pm 0.4}$ & \worse 68.9$_{\pm 0.6}$ & 73.4$_{\pm 0.5}$ & \better 31.7$_{\pm 1.8}$ & \better 61.3$_{\pm 1.2}$ & 33.6$_{\pm 0.5}$ & 29.7$_{\pm 0.2}$ & 52.8$_{\pm 0.7}$ \\
+ Weak Align (Our) & \better 29.9$_{\pm 1.0}$ & 67.6$_{\pm 0.4}$ & \better 79.8$_{\pm 0.0}$ & \better 69.6$_{\pm 0.3}$ & 73.5$_{\pm 0.5}$ & 31.0$_{\pm 1.6}$ & \better 61.2$_{\pm 0.9}$ & 33.4$_{\pm 0.7}$ & \better 30.0$_{\pm 0.5}$ & 52.9$_{\pm 0.6}$ \\
+ Strong Align (Our) & \better 30.8$_{\pm 0.9}$ & 68.0$_{\pm 0.4}$ & \better 79.8$_{\pm 0.1}$ & \better 69.9$_{\pm 0.3}$ & \better 73.7$_{\pm 0.5}$ & \better 31.5$_{\pm 1.5}$ & \better 61.8$_{\pm 0.6}$ & 33.5$_{\pm 0.6}$ & \better 30.4$_{\pm 0.4}$ & \better 53.3$_{\pm 0.6}$ \\

\midrule
XLMR$_\text{base}$ & 43.7$_{\pm 1.7}$ & 69.0$_{\pm 0.4}$ & 80.5$_{\pm 0.2}$ & 71.0$_{\pm 0.4}$ & 73.6$_{\pm 0.5}$ & 41.2$_{\pm 0.9}$ & 66.3$_{\pm 0.9}$ & 36.6$_{\pm 0.2}$ & 34.2$_{\pm 0.7}$ & 57.3$_{\pm 0.6}$ \\
+ Linear Mapping & \better 48.0$_{\pm 0.5}$ & 69.2$_{\pm 0.2}$ & \better 81.4$_{\pm 0.1}$ & \better 72.4$_{\pm 0.1}$ & \better 74.8$_{\pm 0.3}$ & \worse 38.8$_{\pm 0.9}$ & \worse 61.8$_{\pm 0.5}$ & \worse 34.2$_{\pm 0.3}$ & \worse 24.2$_{\pm 0.9}$ & \worse 56.1$_{\pm 0.4}$ \\
+ L2 Align & \worse 39.4$_{\pm 0.5}$ & \worse 68.0$_{\pm 0.5}$ & \worse 79.9$_{\pm 0.2}$ & \worse 69.9$_{\pm 0.5}$ & \worse 72.8$_{\pm 0.5}$ & \worse 40.2$_{\pm 1.1}$ & \worse 63.8$_{\pm 0.8}$ & 36.4$_{\pm 0.6}$ & \worse 32.3$_{\pm 0.9}$ & \worse 55.9$_{\pm 0.6}$ \\
+ Weak Align (Our) & 44.5$_{\pm 1.3}$ & 68.7$_{\pm 0.7}$ & 80.4$_{\pm 0.1}$ & 71.3$_{\pm 0.3}$ & 73.8$_{\pm 0.3}$ & 41.4$_{\pm 0.8}$ & 65.7$_{\pm 0.4}$ & 36.7$_{\pm 0.4}$ & 34.0$_{\pm 0.7}$ & 57.4$_{\pm 0.5}$ \\
+ Strong Align (Our) & 44.9$_{\pm 1.0}$ & 68.8$_{\pm 0.6}$ & 80.4$_{\pm 0.1}$ & 71.2$_{\pm 0.2}$ & 73.8$_{\pm 0.2}$ & 41.1$_{\pm 0.8}$ & 65.9$_{\pm 0.5}$ & 36.6$_{\pm 0.3}$ & 33.9$_{\pm 0.7}$ & 57.4$_{\pm 0.5}$ \\

\midrule
XLMR$_{\text{large}}$ & 48.2$_{\pm 1.5}$ & 67.8$_{\pm 0.6}$ & 82.6$_{\pm 0.3}$ & 73.9$_{\pm 0.4}$ & 76.4$_{\pm 0.4}$ & 41.8$_{\pm 2.5}$ & 69.6$_{\pm 0.4}$ & 38.9$_{\pm 0.6}$ & 35.4$_{\pm 0.5}$ & 59.4$_{\pm 0.8}$ \\

\bottomrule
\end{tabular}
}
\caption{Zero-shot cross-lingual transfer result with the OPUS-100 bitext. 
\textcolor{better}{Blue} or \textcolor{worse}{orange} indicates the mean performance is one standard derivation \textcolor{better}{above} or \textcolor{worse}{below} the mean of baseline.
\label{tab:all-opus}}
\end{center}
\end{table*}
}

\newcommand{\insertAverTable}{
\begin{table*}[t]
\begin{subtable}{.49\linewidth}
\begin{center}
\resizebox{1\linewidth}{!}{
\begin{tabular}[b]{l|cccc}
\toprule
 & XNLI & NER & POS & Parsing \\
\midrule

mBERT & 70.1$_{\pm 0.8}$ & 67.7$_{\pm 1.3}$ & 78.3$_{\pm 0.5}$ & 52.6$_{\pm 0.4}$ \\
+ Linear Mapping & 70.0$_{\pm 0.6}$ & \worse 63.7$_{\pm 1.5}$ & \better 79.5$_{\pm 0.5}$ & \better 53.6$_{\pm 0.3}$ \\
+ L2 Align & 69.7$_{\pm 0.4}$ & 67.1$_{\pm 1.0}$ & 78.0$_{\pm 1.3}$ & 52.2$_{\pm 0.7}$ \\
+ Weak Align (Our) & 70.5$_{\pm 0.7}$ & 68.0$_{\pm 1.3}$ & \better 78.8$_{\pm 0.7}$ & \better 53.1$_{\pm 0.6}$ \\
+ Strong Align (Our) & 70.4$_{\pm 0.7}$ & 67.7$_{\pm 1.1}$ & \better 79.0$_{\pm 0.7}$ & 53.0$_{\pm 0.6}$ \\
\midrule

XLMR$_{\text{base}}$ & 76.4$_{\pm 0.5}$ & 66.4$_{\pm 0.9}$ & 81.2$_{\pm 0.6}$ & 57.3$_{\pm 0.6}$ \\
+ Linear Mapping & \worse 73.4$_{\pm 0.6}$ & \worse 54.1$_{\pm 0.9}$ & 81.3$_{\pm 0.5}$ & \worse 55.6$_{\pm 0.5}$ \\
+ L2 Align & \worse 75.7$_{\pm 0.5}$ & 65.7$_{\pm 1.2}$ & 81.3$_{\pm 0.9}$ & \worse 56.2$_{\pm 0.7}$ \\
+ Weak Align (Our) & 76.1$_{\pm 0.7}$ & 66.0$_{\pm 1.0}$ & 81.5$_{\pm 0.5}$ & 57.4$_{\pm 0.4}$ \\
+ Strong Align (Our) & 76.0$_{\pm 0.6}$ & 66.1$_{\pm 0.9}$ & 81.4$_{\pm 0.6}$ & 57.4$_{\pm 0.5}$ \\
\midrule

XLMR$_{\text{large}}$ & 80.4$_{\pm 0.6}$  & 71.0$_{\pm 1.4}$  & 82.6$_{\pm 0.5}$  & 59.4$_{\pm 0.8}$  \\
\bottomrule
\end{tabular}
}
\caption{Alignment with bitext used in previous works}
\end{center}
\end{subtable}
\begin{subtable}{.49\linewidth}
\begin{center}
\resizebox{1\linewidth}{!}{
\begin{tabular}[b]{l|cccc}
\toprule
 & XNLI & NER & POS & Parsing \\
\midrule

mBERT & 70.1$_{\pm 0.8}$ & 67.7$_{\pm 1.3}$ & 78.3$_{\pm 0.5}$ & 52.6$_{\pm 0.4}$ \\
+ Linear Mapping & 70.2$_{\pm 0.6}$ & \worse 63.8$_{\pm 1.3}$ & \better 80.1$_{\pm 0.4}$ & \better 53.6$_{\pm 0.3}$ \\
+ L2 Align & 70.3$_{\pm 0.5}$ & 67.8$_{\pm 1.4}$ & 78.2$_{\pm 1.2}$ & 52.8$_{\pm 0.7}$ \\
+ Weak Align (Our) & 70.8$_{\pm 0.7}$ & 67.3$_{\pm 0.9}$ & \better 78.8$_{\pm 0.6}$ & 52.9$_{\pm 0.6}$ \\
+ Strong Align (Our) & 70.4$_{\pm 0.7}$ & 67.2$_{\pm 1.1}$ & \better 79.0$_{\pm 0.7}$ & \better 53.3$_{\pm 0.6}$ \\
\midrule

XLMR$_{\text{base}}$ & 76.4$_{\pm 0.5}$ & 66.4$_{\pm 0.9}$ & 81.2$_{\pm 0.6}$ & 57.3$_{\pm 0.6}$ \\
+ Linear Mapping & \worse 73.5$_{\pm 0.5}$ & \worse 54.2$_{\pm 0.8}$ & 81.7$_{\pm 0.6}$ & \worse 56.1$_{\pm 0.4}$ \\
+ L2 Align & \worse 75.8$_{\pm 0.5}$ & \worse 65.5$_{\pm 1.2}$ & 81.4$_{\pm 0.8}$ & \worse 55.9$_{\pm 0.6}$ \\
+ Weak Align (Our) & 76.0$_{\pm 0.4}$ & 66.2$_{\pm 1.2}$ & 81.5$_{\pm 0.5}$ & 57.4$_{\pm 0.5}$ \\
+ Strong Align (Our) & 76.1$_{\pm 0.4}$ & 66.2$_{\pm 1.0}$ & 81.5$_{\pm 0.6}$ & 57.4$_{\pm 0.5}$ \\
\midrule

XLMR$_{\text{large}}$ & 80.4$_{\pm 0.6}$  & 71.0$_{\pm 1.4}$  & 82.6$_{\pm 0.5}$  & 59.4$_{\pm 0.8}$  \\
\bottomrule
\end{tabular}
}
\caption{Alignment with the OPUS-100 bitext}
\end{center}
\end{subtable}

\caption{Zero-shot cross-lingual transfer result, average over 9 languages. Breakdown can be found in \cref{sec:breakdown}.
\textcolor{better}{Blue} or \textcolor{worse}{orange} indicates the mean performance is one standard derivation \textcolor{better}{above} or \textcolor{worse}{below} the mean of baseline. While mBERT benefits from alignment in some cases, extra alignment does not improve XLMR.\label{tab:aver}}
\end{table*}
}

\newcommand{\insertStatTable}{
\begin{table}[h]
\begin{center}
\resizebox{0.8\linewidth}{!}{
\begin{tabular}[b]{cccc}
\toprule
 & \multirow{2}{*}{XNLI} & \multirow{2}{*}{NER} & POS tagging \\
 & & & Parsing\\
\midrule
en-train & 392703 & 20000 & 12543 \\
en-dev & 2490 & 10000 & 2002 \\
en-test & 5010 & 10000 & 2077 \\
\midrule
ar-test & 5010 & 10000 & 680 \\
de-test & 5010 & 10000 & 977 \\
es-test & 5010 & 10000 & 426 \\
fr-test & 5010 & 10000 & 416 \\
hi-test & 5010 & 1000 & 1684 \\
ru-test & 5010 & 10000 & 601 \\
vi-test & 5010 & 10000 & 800 \\
zh-test & 5010 & 10000 & 500 \\
\bottomrule
\end{tabular}
}
\caption{Number of examples.\label{tab:stat}}
\end{center}
\vspace{-0.4cm}
\end{table}
}